\documentclass{article}

    \PassOptionsToPackage{numbers, compress}{natbib}

\usepackage[main, final]{neurips_2025}

\usepackage[utf8]{inputenc} 
\usepackage[T1]{fontenc}    
\usepackage{hyperref}       
\usepackage{url}            
\usepackage{booktabs}       
\usepackage{amsfonts}       
\usepackage{nicefrac}       
\usepackage{microtype}      
\usepackage{xcolor}         
 \usepackage[pdftex]{graphicx}
\title{A tutorial on discovering and quantifying the effect of latent causal sources of multimodal EHR data}

\author{
  Marco Barbero-Mota\thanks{\texttt{marco.barbero.mota@vanderbilt.edu}} \\
  Department of Biomedical Informatics\\
  Vanderbilt University Medical Center\\
  \\
  \And
  Eric V. Strobl \\
  Department of Biomedical Informatics\\
  University of Pittsburgh \\
  \AND
  John M. Still \\
  Department of Biomedical Informatics \\
  Vanderbilt University Medical Center\\
  \And
  William W. Stead \\
  Departments of Medicine \& Biomedical Informatics\\
  Vanderbilt University Medical Center\\
  \And
 Thomas A. Lasko \\
  Departments of Biomedical Informatics \& Computer Science\\
  Vanderbilt University Medical Center \& Vanderbilt University\\
}

\begin{document}

\maketitle

\begin{abstract}
We provide an accessible description of a peer-reviewed generalizable causal machine learning pipeline to (i) discover latent causal sources of large-scale electronic health records observations, and (ii) quantify the source causal effects on clinical outcomes. We illustrate how imperfect multimodal clinical data can be processed, decomposed into probabilistic independent latent sources, and used to train task-specific causal models from which individual causal effects can be estimated. We summarize the findings of the two real-world applications of the approach to date as a demonstration of its versatility and utility for medical discovery at scale.
\end{abstract}

\section{Introduction}
A central objective of biomedical research is to identify the causes of disease. Causal analyses seek to disentangle the effects of interventions and apply this knowledge to guide decisions that improve patient outcomes. However, estimating causal effects for individual patients remains challenging, in part due to latent confounding. Randomized clinical trials address confounding by assigning treatments randomly and therefore remain the gold standard for causal effect estimation. Nevertheless, clinical trials typically investigate only a limited set of interventions (e.g., treatment \textit{vs}. placebo), apply conservative inclusion criteria, and enroll relatively small samples because of their high cost. Consequently, they often provide sufficient data only to estimate aggregate causal effects, such as the average treatment effect, rather than individualized effects conditional on patient covariates, and may suffer from selection bias and limited generalizability \citep{averitt2020translating, rothwell2006factors, deaton2018understanding}.

We can attempt to overcome the sample size limitations of clinical trials by approximating randomized experiments with observational data, provided that we condition on the appropriate covariates to mitigate confounding and other biases. The key challenge, however, is identifying which covariates to adjust for. Causal machine learning (ML) models provide a promising framework by enabling proper adjustment even for large covariate sets derived from multimodal Electronic Health Records (EHRs) \citep{feuerriegel2024causal, doutreligne2025step}. Nonetheless, because EHR data are collected primarily for clinical care rather than research, they remain subject to latent confounding, implicit selection biases, and other systemic errors, such as  process-related biases, that hinder causal effect identification despite large adjustment sets \citep{zhang2019medical, agniel2018biases, lasko2024probabilistic} . Moreover, EHR data introduce additional practical challenges, including systematic and random noise, longitudinal sparsity and incompleteness. 

EHR data nevertheless also provide \textit{advantages} that extend beyond the capabilities of randomized trials and other observational datasets. The EHR reflects aggregated practical medical knowledge, captures key aspects of patient physiology, and implicitly records how care is delivered and documented \citep{hripcsak2013next}. Clinicians can thus routinely incorporate information from the EHR into their reasoning to infer likely causes of disease compatible with observed clinical states \citep{strobl2024counterfactual}, and to evaluate potential interventions most likely to address those causes and improve patient outcomes \citep{strobl2022identifying, feuerriegel2024causal}. We thus ask:

\textbf{Can we computationally infer latent causes from their effects in the EHR and quantify their individual effects at scale?} We argue that the answer is affirmative, though it requires both caution and nuance. Many existing causal ML methods rely on strong untestable assumptions, raising uncertainty about their causal validity when directly applied to large imperfect EHR data \cite{pearl2009causality,kaddour2022causal}. To address this challenge, we created this tutorial to provide a clear, step-by-step summary of the data-driven pipeline introduced in \citet{strobl2022identifying} and \citet{strobl2024counterfactual}, and later scaled to large multimodal EHR data in \citet{lasko2025unsupervised}. We demonstrate that the pipeline (i) addresses many of the intrinsic challenges of processing large EHR datasets; (ii) infers probabilistically independent latent sources that act as candidate causes of EHR patterns in high-dimensional settings; and (iii) quantifies the causal effects of these inferred sources on a patient and timepoint-level for specific outcomes. Note that (ii) and (iii) parallel  Steps 1 and 3 of Pearl's counterfactual inference \citep{pearl2009causality}. Crucially, the pipeline circumvents the difficult problem of estimating other intermediary causal structures often required for effect estimation in practice, such as causal graphs or structural equations of the underlying causal processes \cite{pearl2009causality,strobl2022identifying}. We further illustrate its potential through two peer-reviewed real-world applications, framed as predictive tasks in distinct health conditions.

\begin{figure}
    \centering
    \hspace{-1cm}\includegraphics[width=0.9\linewidth]{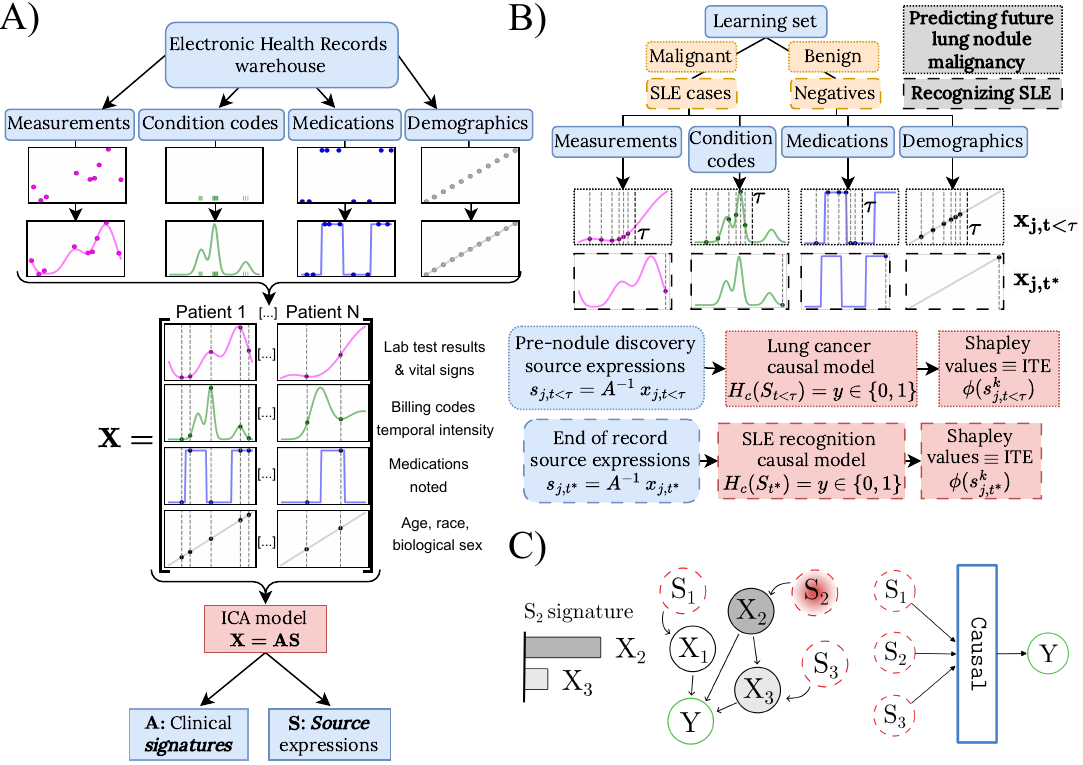}
    \caption{Causal ML pipeline summary. Information flows from multimodal episodic EHR observations that are preprocessed for \textbf{A)} independent latent causal sources \emph{discovery}, to \textbf{B)} task-specific supervised causal models $H_c(S)$. $H_c$ learns to predict $Y$ from the source expressions of a small labeled cohort. Sample-specific Shapley additive explanations $\phi$ estimate individual treatment effects at the $(j,t)$ patient-timepoint level for each of $H_c(S)$ $k$ input sources. \textbf{C)}  The source inputs to $H_c(S)$ are latent exogenous nodes of the underlying linear SCM causal graph. When active, predictive sources leave a \emph{signature} in the observed variables and ultimately affect $Y$.}
    \label{fig:summary}
\end{figure}

\section{Preparing EHR data for machine learning}
EHRs provide rich, high-dimensional multimodal data particularly well-suited for ML-based knowledge discovery. However, their episodic nature requires substantial preprocessing to transform sparse, noisy and incomplete observations into a substrate suitable for ML methods. Prior work \citep{lasko2013computational, lasko2019computational, lasko2025unsupervised} developed and progressively refined a modular pipeline that can process institution-wide datasets with thousands of variables and a large proportion of missing values into a standardized and complete data matrix ready for downstream analyses. The constituent steps are illustrated in Figure \ref{fig:summary} A.

First, we ingest patient visit data for a large unlabeled discovery cohort over four structured data modalities: \textit{measurements}, such as laboratory test results and vital signs; \emph{condition codes} for diagnoses, also referred to as billing codes since their main purpose is to document the complexity of a visit for reimbursement purposes; mentions of \textit{medications}; and \textit{demographics} including race, biological sex and age. Unstructured data modalities such as medical images or clinical notes could also be included by, for example, extracting tabular features from them.  

Next, we infer continuous daily-resolution trajectories for each patient variable using modality-specific, clinically-informed methods. This step attempts to overcome within-record data missingness and asynchronicity across clinical variables. \emph{Measurement} curves are smooth piecewise cubic interpolations (PCHIP) \cite{fritsch1984method} that avoid overshooting and maintain monotonicity and nonstationarity where present. \emph{Condition codes} are transformed into intensity curves (i.e. instantaneous code frequency per unit time) using a modified version of Random Average Shifted Histograms (RASH) \cite{bourel2014random} that accommodates nonstationarity in event-arrival density over a record timespan. \emph{Medication} mentions are converted into binary curves that approximate taking \textit{vs.} not taking by extrapolation from the nearest reconciliation dates. \emph{Demographics} curves are constant one-hot encoded for categorical variables (race and biological sex) and linearly increasing for age. Completely missing \emph{measurements}, \emph{condition codes}, and \emph{medications} in a given record are imputed with population median, a baseline intensity prior of one code per 20 years, and constant 0 (not taking) curves, respectively.

For each patient, the curves are time-aligned into a curveset from which we sample cross sections at a number of random times proportional to the record length and the desired sampling density. Cross sections are then stacked into a dense input matrix $X$ that is then standardized to put all variables on roughly the same scale. For complete data processing details refer to Appendix A in \citet{lasko2025unsupervised}.

\section{Discovering causal \emph{sources} and their EHR \emph{signatures}}
The matrix $X$ forms the data input to the \emph{discovery step}, with rows representing the EHR variables and columns the record cross-sections. We use the FastICA algorithm to learn an independent component analysis (ICA) model from $X$. ICA finds a linear unmixing of its input into probabilistically independent latent \emph{sources}. All sources except one are assumed to be non-gaussian. ICA therefore decomposes $X$ into a mixing matrix $A$ and a source matrix $S$ such that $X=AS$ \cite{hyvarinen_independent_2000, hyvarinen2001independent}. Rows of $S$ correspond to the mutually independent latent sources, and columns to the cross sections in the corresponding columns of $X$. The values in $S$ represent the level at which each source is expressed at each cross section. We refer to the columns of $A$ as source \emph{signatures} representing the linear changes in the original data space caused by one unit of expression of a given source. For EHR data, the \textit{i-th} column of $A$ specifies the changes that a one unit change in $S_{i,:}$ imprints to the record across all observed clinical variables.

From Pearl's perspective, a data generation process (DGP) can be formalized through a structural causal model (SCM). Each SCM has an associated directed acyclic graph (Figure \ref{fig:summary} C) and a set of equations that encode the functional relationships among exogenous latent sources and endogenous observed variables \citep{pearl2009causality}. \citet{shimizu2006linear} first described the connection between ICA and linear SCM equations. The identifiability of ICA allows unique recovery of an SCM's exogenous sources, given observations of the endogenous variables, under the linear acyclic non-Gaussian model assumptions \citep{shimizu2006linear, hyvarinen2024identifiability}. 

In this context, \citet{strobl2022identifying} rigorously defined the sources as \emph{root causes of disease} given observed data. Note that the root level of ICA discovered sources goes as far back in the complete causal graph as the input data allows. For most diseases and EHR datasets, information from the true root causes, such as environmental or genetic factors, is usually upstream from the observed variables in $X$, and only their effects can be seen in the signatures in $A$. Each source can reflect pathophysiological pathways, clinical workflow patterns, or a mixture of these \cite{lasko2024probabilistic}. Expert interpretation of a source signature can help identify its clinical meaning and validation, as we will illustrate below.

\section{Causally predictive models}
Our objective is to find the subset of sources in $S$ that are causally predictive of a health outcome of interest $Y$. We formulate the problem as a predictive task, where we optimize a supervised architecture to learn to predict $Y$ from source expressions of a labeled patient cohort: $H_c(S) = Y$. Under a weaker unconfoundedness assumption than usual \citep{lasko2025unsupervised, wang2019blessings}, and the assumption that $Y$ is a sink node in the prediction problem causal graph \citep{strobl2022identifying}, learning from parentless independent nodes constrains the problem such that the only flow of information between the model predictors and $Y$ is through causal paths (Figure \ref{fig:summary}C). This is in contrast to statistical learning in the original data space ($H_a(X) = Y$), which is susceptible to finding highly predictive but non-causal shortcuts in-sample when minimizing prediction error.
In theory, causal identification intrinsic to the sources allows using any supervised learning architecture to learn a descriptive model of $Y$ which confers greater interpretability to clinical ML models.

Given a trained causal model $H_c(S)$, \citet{strobl2022identifying} identified sample-specific Shapley additive explanation (SHAP) values \citep{lundberg_unified_2017, LundbergFromTrees} as the mathematical formulation for quantifying each source's total causal effect on $Y$ at the instance level; i.e., each source's individual treatment effects (ITE). Although SHAP values are usually employed as an associational measure of feature importance, when computed over the causal model $H_c(S)$, independence among predictors moves up SHAP values from rung 1 in Pearl's ladder of causation, thereby quantifying causal effects \citep{strobl2022identifying, strobl2024counterfactual, pearl2009causality}. $H_c(S)$ SHAP values estimate each source's attributable impact on $Y$ with respect to a reference "healthy" value, technically referred to as the base value. A source SHAP value captures both its marginal and joint effects from all possible combinations with other sources \citep{strobl2022identifying, janzing2020feature}.

\section{Real-world examples}

\textbf{Lung cancer [Figure \ref{fig:summary}C]}. \citet{lasko2025unsupervised} applied the above pipeline to discover 2000 latent causal sources of EHR data for all institution-wide patients with lung disease, defined broadly. From that set of latent causes the authors identified those predictive of \emph{future} malignancy among patients diagnosed with indeterminate pulmonary nodules and no prior history of any cancer. The analysis recovered $92\%$ of established causes of malignant lung cancer. Furthermore, of the top 20 most predictive sources, 6 ($30\%$) were recognized etiologies, 10 ($50\%$) had various level of empirical support in the pulmonary oncology literature, and 4 ($20\%$) had no explicit peer-reviewed evidence but enough clinical face validity to grant further investigation. 
    
\textbf{Systemic Lupus Erythematosus (SLE) [Figure \ref{fig:summary}B]}. \citet{mota2025data} performed the same analysis of EHR latent causes for patients who had an antinuclear antibody test done, presumably under the suspicion of autoimmune disease. A causal model was trained to recognize SLE in the health record, a task also known as computational phenotyping \citep{banda2018advances}. Predictive sources of SLE chart review labels included multiple disease manifestations: two canonical presentations varying in the specific SLE billing code used, recognizable SLE components (lupus nephritis and four antiphospholipid syndrome forms), a laboratory biomarker of disease activity, SLE treatments, overlapping diseases (rheumatoid arthritis and Sjögren's syndrome), and a surprising operational source capturing a secondary effect of rheumatic autoimmune disorders. Its signature revealed an EHR pattern that clinicians leave when prospectively coding for a rare eye condition, toxic maculopathy, as insurance justification for retinal screening in hydroxychloroquine long‑term users. Toxic maculopathy codes are an unconventional predictor of the SLE label, but causal ML was able to identify this process-based latent source. 

Results of both studies highlight (i) the value of \textbf{causal ML guided by probabilistic independence} as an etiological hypothesis generation at scale, and (ii) its broad applicability across medical specialties. Importantly, both evaluations found no significant in-distribution performance boost for causal models over their correlational counterpart. However, the main objective of causal modeling is not to minimize in-sample predictive error, but rather to find stable predictive descriptors of reality \citep{arjovsky2019invariant}, produce higher model interpretability \citep{lipton2018mythos}, and enable data-driven discovery \citep{glymour2019review}.

\textbf{Future directions}. Currently, our descriptive causal models are trained to minimize the predictive error of $Y = H_c(S)$.  However, the loss functions for optimal prediction and optimal inference of internal causal structure are unlikely to coincide. How to choose the combination of hypothesis space, loss function, and optimization approach to maximize causal structure accuracy is an important but unexplored area of research.

A second important and unsolved problem is evaluating causal discovery results, because we lack ground truth for both the identification of latent sources and any counterfactual implications of the model structure \cite{alaa2019validating, parikh2022validating}. In our work to date, we use domain knowledge to validate inferred causes \cite{lasko2025unsupervised} in an adaptation of quantitative probing, \citep{grunbaum2023quantitative} but this does not validate unknown disease causes.   

\newpage
\bibliographystyle{plainnat}
\bibliography{references}

\begin{thebibliography}{34}
\providecommand{\natexlab}[1]{#1}
\providecommand{\url}[1]{\texttt{#1}}
\expandafter\ifx\csname urlstyle\endcsname\relax
  \providecommand{\doi}[1]{doi: #1}\else
  \providecommand{\doi}{doi: \begingroup \urlstyle{rm}\Url}\fi

\bibitem[Agniel et~al.(2018)Agniel, Kohane, and Weber]{agniel2018biases}
Denis Agniel, Isaac~S Kohane, and Griffin~M Weber.
\newblock Biases in electronic health record data due to processes within the healthcare system: retrospective observational study.
\newblock \emph{Bmj}, 361, 2018.

\bibitem[Alaa and Van Der~Schaar(2019)]{alaa2019validating}
Ahmed Alaa and Mihaela Van Der~Schaar.
\newblock Validating causal inference models via influence functions.
\newblock In \emph{International Conference on Machine Learning}, pages 191--201. PMLR, 2019.

\bibitem[Arjovsky et~al.(2019)Arjovsky, Bottou, Gulrajani, and Lopez-Paz]{arjovsky2019invariant}
Martin Arjovsky, L{\'e}on Bottou, Ishaan Gulrajani, and David Lopez-Paz.
\newblock Invariant risk minimization.
\newblock \emph{arXiv preprint arXiv:1907.02893}, 2019.

\bibitem[Averitt et~al.(2020)Averitt, Weng, Ryan, and Perotte]{averitt2020translating}
Amelia~J Averitt, Chunhua Weng, Patrick Ryan, and Adler Perotte.
\newblock Translating evidence into practice: eligibility criteria fail to eliminate clinically significant differences between real-world and study populations.
\newblock \emph{NPJ digital medicine}, 3\penalty0 (1):\penalty0 67, 2020.

\bibitem[Banda et~al.(2018)Banda, Seneviratne, Hernandez-Boussard, and Shah]{banda2018advances}
Juan~M Banda, Martin Seneviratne, Tina Hernandez-Boussard, and Nigam~H Shah.
\newblock Advances in electronic phenotyping: from rule-based definitions to machine learning models.
\newblock \emph{Annual review of biomedical data science}, 1\penalty0 (1):\penalty0 53--68, 2018.

\bibitem[Bourel et~al.(2014)Bourel, Fraiman, and Ghattas]{bourel2014random}
Mathias Bourel, Ricardo Fraiman, and Badih Ghattas.
\newblock Random average shifted histograms.
\newblock \emph{Computational Statistics \& Data Analysis}, 79:\penalty0 149--164, 2014.

\bibitem[Deaton and Cartwright(2018)]{deaton2018understanding}
Angus Deaton and Nancy Cartwright.
\newblock Understanding and misunderstanding randomized controlled trials.
\newblock \emph{Social science \& medicine}, 210:\penalty0 2--21, 2018.

\bibitem[Doutreligne et~al.(2025)Doutreligne, Struja, Abecassis, Morgand, Celi, and Varoquaux]{doutreligne2025step}
Matthieu Doutreligne, Tristan Struja, Judith Abecassis, Claire Morgand, Leo~Anthony Celi, and Ga{\"e}l Varoquaux.
\newblock Step-by-step causal analysis of ehrs to ground decision-making.
\newblock \emph{PLOS Digital Health}, 4\penalty0 (2):\penalty0 e0000721, 2025.

\bibitem[Feuerriegel et~al.(2024)Feuerriegel, Frauen, Melnychuk, Schweisthal, Hess, Curth, Bauer, Kilbertus, Kohane, and van~der Schaar]{feuerriegel2024causal}
Stefan Feuerriegel, Dennis Frauen, Valentyn Melnychuk, Jonas Schweisthal, Konstantin Hess, Alicia Curth, Stefan Bauer, Niki Kilbertus, Isaac~S Kohane, and Mihaela van~der Schaar.
\newblock Causal machine learning for predicting treatment outcomes.
\newblock \emph{Nature Medicine}, 30\penalty0 (4):\penalty0 958--968, 2024.

\bibitem[Fritsch and Butland(1984)]{fritsch1984method}
Frederick~N Fritsch and Judy Butland.
\newblock A method for constructing local monotone piecewise cubic interpolants.
\newblock \emph{SIAM journal on scientific and statistical computing}, 5\penalty0 (2):\penalty0 300--304, 1984.

\bibitem[Glymour et~al.(2019)Glymour, Zhang, and Spirtes]{glymour2019review}
Clark Glymour, Kun Zhang, and Peter Spirtes.
\newblock Review of causal discovery methods based on graphical models.
\newblock \emph{Frontiers in genetics}, 10:\penalty0 524, 2019.

\bibitem[Gr{\"u}nbaum et~al.(2023)Gr{\"u}nbaum, Stern, and Lang]{grunbaum2023quantitative}
Daniel Gr{\"u}nbaum, Maike~L Stern, and Elmar~W Lang.
\newblock Quantitative probing: Validating causal models with quantitative domain knowledge.
\newblock \emph{Journal of Causal Inference}, 11\penalty0 (1):\penalty0 20220060, 2023.

\bibitem[Hripcsak and Albers(2013)]{hripcsak2013next}
George Hripcsak and David~J Albers.
\newblock Next-generation phenotyping of electronic health records.
\newblock \emph{J Am Med Inform Assoc}, 20:\penalty0 117--121, 2013.

\bibitem[Hyv{\"a}rinen et~al.(2001)Hyv{\"a}rinen, Hurri, and Hoyer]{hyvarinen2001independent}
Aapo Hyv{\"a}rinen, Jarmo Hurri, and Patrik~O Hoyer.
\newblock Independent component analysis.
\newblock In \emph{Natural Image Statistics: A Probabilistic Approach to Early Computational Vision}, pages 151--175. Springer, 2001.

\bibitem[Hyv{\"a}rinen et~al.(2024)Hyv{\"a}rinen, Khemakhem, and Monti]{hyvarinen2024identifiability}
Aapo Hyv{\"a}rinen, Ilyes Khemakhem, and Ricardo Monti.
\newblock Identifiability of latent-variable and structural-equation models: from linear to nonlinear.
\newblock \emph{Annals of the Institute of Statistical Mathematics}, 76\penalty0 (1):\penalty0 1--33, 2024.

\bibitem[Hyvärinen and Oja(2000)]{hyvarinen_independent_2000}
Aapo Hyvärinen and Erkki Oja.
\newblock Independent {Component} {Analysis}: {Algorithms} and {Applications}.
\newblock \emph{Neural Networks}, 13\penalty0 (5):\penalty0 411--430, 2000.

\bibitem[Janzing et~al.(2020)Janzing, Minorics, and Bl{\"o}baum]{janzing2020feature}
Dominik Janzing, Lenon Minorics, and Patrick Bl{\"o}baum.
\newblock Feature relevance quantification in explainable ai: A causal problem.
\newblock In \emph{International Conference on artificial intelligence and statistics}, pages 2907--2916. PMLR, 2020.

\bibitem[Kaddour et~al.(2022)Kaddour, Lynch, Liu, Kusner, and Silva]{kaddour2022causal}
Jean Kaddour, Aengus Lynch, Qi~Liu, Matt~J Kusner, and Ricardo Silva.
\newblock Causal machine learning: A survey and open problems.
\newblock \emph{arXiv preprint arXiv:2206.15475}, 2022.

\bibitem[Lasko and Mesa(2019)]{lasko2019computational}
Thomas~A Lasko and Diego~A Mesa.
\newblock Computational phenotype discovery via probabilistic independence.
\newblock In \emph{KDD Workshop on Applied Data Science for Healthcare 2019}, 2019.

\bibitem[Lasko et~al.(2013)Lasko, Denny, and Levy]{lasko2013computational}
Thomas~A Lasko, Joshua~C Denny, and Mia~A Levy.
\newblock Computational phenotype discovery using unsupervised feature learning over noisy, sparse, and irregular clinical data.
\newblock \emph{PloS one}, 8\penalty0 (6):\penalty0 e66341, 2013.

\bibitem[Lasko et~al.(2024)Lasko, Strobl, and Stead]{lasko2024probabilistic}
Thomas~A Lasko, Eric~V Strobl, and William~W Stead.
\newblock Why do probabilistic clinical models fail to transport between sites.
\newblock \emph{NPJ Digital Medicine}, 7\penalty0 (1):\penalty0 53, 2024.

\bibitem[Lasko et~al.(2025)Lasko, Stead, Still, Li, Kammer, Barbero-Mota, Strobl, Landman, and Maldonado]{lasko2025unsupervised}
Thomas~A Lasko, William~W Stead, John~M Still, Thomas~Z Li, Michael Kammer, Marco Barbero-Mota, Eric~V Strobl, Bennett~A Landman, and Fabien Maldonado.
\newblock Unsupervised discovery of clinical disease signatures using probabilistic independence.
\newblock \emph{Journal of Biomedical Informatics}, page 104837, 2025.

\bibitem[Lipton(2018)]{lipton2018mythos}
Zachary~C Lipton.
\newblock The mythos of model interpretability: In machine learning, the concept of interpretability is both important and slippery.
\newblock \emph{Queue}, 16\penalty0 (3):\penalty0 31--57, 2018.

\bibitem[Lundberg et~al.()Lundberg, Erion, Chen, Degrave, Prutkin, Nair, Katz, Himmelfarb, Bansal, and Lee]{LundbergFromTrees}
Scott~M Lundberg, Gabriel Erion, Hugh Chen, Alex Degrave, Jordan~M Prutkin, Bala Nair, Ronit Katz, Jonathan Himmelfarb, Nisha Bansal, and Su-In Lee.
\newblock {From local explanations to global understanding with explainable AI for trees}.
\newblock \emph{Nature Machine Intelligence}.
\newblock \doi{10.1038/s42256-019-0138-9}.
\newblock URL \url{https://doi.org/10.1038/s42256-019-0138-9}.

\bibitem[Lundberg et~al.(2017)Lundberg, Allen, and Lee]{lundberg_unified_2017}
Scott~M Lundberg, Paul~G Allen, and Su-In Lee.
\newblock A {Unified} {Approach} to {Interpreting} {Model} {Predictions}.
\newblock \emph{Advances in Neural Information Processing Systems}, 30, 2017.
\newblock URL \url{https://github.com/slundberg/shap}.

\bibitem[Mota et~al.(2025)Mota, Still, Gamboa, Strobl, Stein, Kawai, and Lasko]{mota2025data}
Marco~Barbero Mota, John~M Still, Jorge~L Gamboa, Eric~V Strobl, Charles~M Stein, Vivian~K Kawai, and Thomas~A Lasko.
\newblock A data-driven approach to discover and quantify systemic lupus erythematosus etiological heterogeneity from electronic health records.
\newblock In \emph{AMIA Annual Symposium Proceedings}, volume 2024, page 172, 2025.

\bibitem[Parikh et~al.(2022)Parikh, Varjao, Xu, and Tchetgen]{parikh2022validating}
Harsh Parikh, Carlos Varjao, Louise Xu, and Eric~Tchetgen Tchetgen.
\newblock Validating causal inference methods.
\newblock In \emph{International conference on machine learning}, pages 17346--17358. PMLR, 2022.

\bibitem[Pearl(2009)]{pearl2009causality}
Judea Pearl.
\newblock \emph{Causality}.
\newblock Cambridge university press, 2009.

\bibitem[Rothwell(2006)]{rothwell2006factors}
Peter~M Rothwell.
\newblock Factors that can affect the external validity of randomised controlled trials.
\newblock \emph{PLoS clinical trials}, 1\penalty0 (1):\penalty0 e9, 2006.

\bibitem[Shimizu et~al.(2006)Shimizu, Hoyer, Hyv{\"a}rinen, Kerminen, and Jordan]{shimizu2006linear}
Shohei Shimizu, Patrik~O Hoyer, Aapo Hyv{\"a}rinen, Antti Kerminen, and Michael Jordan.
\newblock A linear non-gaussian acyclic model for causal discovery.
\newblock \emph{Journal of Machine Learning Research}, 7\penalty0 (10), 2006.

\bibitem[Strobl(2024)]{strobl2024counterfactual}
Eric~V Strobl.
\newblock Counterfactual formulation of patient-specific root causes of disease.
\newblock \emph{Journal of Biomedical Informatics}, 150:\penalty0 104585, 2024.

\bibitem[Strobl and Lasko(2022)]{strobl2022identifying}
Eric~V Strobl and Thomas~A Lasko.
\newblock Identifying patient-specific root causes of disease.
\newblock In \emph{Proceedings of the 13th ACM International Conference on Bioinformatics, Computational Biology and Health Informatics}, pages 1--10, 2022.

\bibitem[Wang and Blei(2019)]{wang2019blessings}
Yixin Wang and David~M Blei.
\newblock The blessings of multiple causes.
\newblock \emph{Journal of the American Statistical Association}, 114\penalty0 (528):\penalty0 1574--1596, 2019.

\bibitem[Zhang et~al.(2019)Zhang, Wang, Ostropolets, Mulgrave, Blei, and Hripcsak]{zhang2019medical}
Linying Zhang, Yixin Wang, Anna Ostropolets, Jami~J Mulgrave, David~M Blei, and George Hripcsak.
\newblock The medical deconfounder: assessing treatment effects with electronic health records.
\newblock In \emph{Machine Learning for Healthcare Conference}, pages 490--512. PMLR, 2019.

\end{thebibliography}

\end{document}